\documentclass[11pt]{article}

\usepackage[T1]{fontenc}
\usepackage[utf8]{inputenc}
\usepackage{lmodern}
\usepackage{microtype}
\usepackage{geometry}
\usepackage{amsmath,amssymb,amsfonts,mathtools}
\usepackage{booktabs}
\usepackage{multirow}
\usepackage{graphicx}
\usepackage{xcolor}
\usepackage{listings}
\usepackage[most]{tcolorbox}
\tcbuselibrary{listings,breakable}
\usepackage{hyperref}
\usepackage[nameinlink,noabbrev]{cleveref}
\usepackage{algorithm}
\usepackage{algpseudocode}
\usepackage{bm} 

\hypersetup{
  colorlinks=true,
  linkcolor=blue,
  citecolor=blue,
  urlcolor=blue
}

\newcommand{\dataset}{ESL-Bench}

\newcommand{\dow}{\mathrm{dow}}

\newtcblisting{promptblock}{
  listing only,
  breakable,
  colback=gray!8,
  colframe=gray!40,
  boxrule=0.4pt,
  arc=2pt,
  left=6pt,right=6pt,top=6pt,bottom=6pt,
  listing options={
    basicstyle=\ttfamily\small,
    columns=fullflexible,
    breaklines=true,
    keepspaces=true,
    showstringspaces=false
  }
}

\title{\dataset{}: An Event-Driven Synthetic Longitudinal Benchmark for Health Agents}
\author{
  Chao Li$^{*}$ \quad
  Cailiang Liu$^{*}$ \quad
  Ang Gao \quad
  Kexin Deng \quad
  Shu Zhang \quad
  Langping Xu \\
  Xiaotong Shi \quad
  Xionghao Ding \quad
  Jian Pei $^{\dagger}$ \quad
  Xun Jiang $^{\dagger}$ \\[4pt]
  {Shanda Group} \\
  \texttt{\{chao.li,cailiang.liu\}@thetahealth.ai} \quad \\
  \texttt{\{ang.gao,kexin.deng,shu.zhang,xulangping\}@thetahealth.ai} \quad \\
  \texttt{\{xiaotong.shi,xionghao.ding\}@thetahealth.ai} \quad \\
  \texttt{j.pei@duke.edu} \quad 
  \texttt{jiangxun@shanda.com}
}
\date{}

\begin{document}
\maketitle
\footnotetext[1]{\textsuperscript{*}These authors contributed equally to this work.}
\footnotetext[2]{\textsuperscript{$\dagger$}Corresponding author.}
\begin{abstract}
Longitudinal health agents must reason across multi-source trajectories
that combine continuous device streams, sparse clinical exams, and
episodic life events---yet evaluating them is hard: real-world data
cannot be released at scale, and temporally grounded attribution
questions seldom admit definitive answers without structured ground truth.
We present \dataset{}, an event-driven synthesis framework and benchmark
providing 100 synthetic users, each with a 1--5~year trajectory
comprising a health profile, a multi-phase narrative plan, daily device
measurements, periodic exam records, and an event log with explicit
per-indicator impact parameters.
Each indicator follows a baseline stochastic process driven by discrete
events with sigmoid-onset, exponential-decay kernels under saturation
and projection constraints; a hybrid pipeline delegates sparse semantic
artifacts to LLM-based planning and dense indicator dynamics to
algorithmic simulation with hard physiological bounds.
Users are each paired with 100 evaluation queries across five
dimensions---Lookup, Trend, Comparison, Anomaly, Explanation---stratified
into Easy, Medium, and Hard tiers, with all ground-truth answers
programmatically computable from the recorded event--indicator
relationships.
Evaluating 13 methods spanning LLMs with tools, DB-native agents, and
memory-augmented RAG, we find that DB agents (48--58\%) substantially
outperform memory RAG baselines (30--38\%), with the gap concentrated
on Comparison and Explanation queries where multi-hop reasoning and
evidence attribution are required.
\end{abstract}

\section{Introduction}
\label{sec:introduction}

\paragraph{Why longitudinal health agents?}
Health data is increasingly longitudinal, heterogeneous, and
patient-generated.
Wearables now stream heart rate, sleep, and activity at daily or
sub-daily cadence outside clinical
walls~\cite{popdig2024,piwek2016wearables,steinhubl2015mhealth};
electronic health records add sparse but high-value clinical snapshots
across visits, labs, and medications.
Making sense of these interleaved streams over months or years is the job of
\emph{longitudinal health agents}---systems that ingest multi-year patient
trajectories to answer questions and ground their answers in verifiable
evidence.

\paragraph{The core challenge: structured temporal reasoning.}
Single-note question answering is not enough.
Longitudinal workloads demand \emph{temporal and relational operations}:
date alignment and unit normalization across sources;
aggregation over event-aligned windows (pre-, during-, and post-event);
multi-hop joins linking events, indicators, and exams; and
attribution-style reasoning that traces observed changes back to plausible
drivers through verifiable evidence chains.
This is a \emph{structured temporal reasoning} problem, not merely a
long-context one, and its failure modes are hard to surface under weak
evaluation~\cite{ehrshot2023,emrqa2018,lee2022ehrsql,medagentbench2025}.

\paragraph{Existing datasets fall short.}
Two barriers block reproducible evaluation: limited data access and the
absence of answerable ground truth for temporally grounded questions.
MIMIC-IV covers acute care but lacks wearable-style device
streams~\cite{mimiciv2022}.
EHRSHOT extends temporal coverage; access is restricted, and the focus is
predictive modeling, not interactive agent
evaluation~\cite{ehrshot2023}.
Clinical QA benchmarks (emrQA~\cite{emrqa2018}, EHRSQL~\cite{lee2022ehrsql},
TIMER~\cite{timer2025}) target specific capabilities---yet attribution in
real EHR data is inherently ambiguous: confounders, missing context, and
overlapping interventions prevent ``what caused this change?'' questions from
admitting definitive answers.
MedAgentBench, MedJourney, and AgentEHR emphasize interactive tool use in EHR
environments but do not test multi-year, multi-source temporal
reasoning~\cite{medagentbench2025,medjourney2024,agentehr2026}.
Meanwhile, distributing multi-year multimodal cohorts at scale remains
prohibitive due to privacy and de-identification
costs~\cite{vanderdonckt2024wearable_data_quality,cho2021pghd_quality}.
No existing benchmark is at once \emph{scalable}, \emph{reproducible}, and
\emph{diagnostic} for agent design choices.

\paragraph{Synthetic data approaches leave a gap.}
Synthetic data offers a path forward, but existing generators sacrifice at
least one of \emph{temporal coherence}, \emph{plausibility}, and
\emph{interpretability}.
Synthea provides explicit clinical logic without continuous, event-aligned
indicator dynamics~\cite{walonoski2018synthea}.
TimeGAN produces realistic-looking sequences, but its dynamics are
opaque---tracing \emph{why} an indicator changed at a given time is
difficult---and plausibility constraints enter only
post-hoc~\cite{yoon2019timegan}.
SynTEG preserves timestamped diagnostic sequences and incorporates privacy
evaluation; it does not, however, decompose trajectories into interpretable
baseline-plus-event contributions~\cite{synteg2021}.
Newer longitudinal synthesis models emphasize distributional fidelity and
downstream
utility~\cite{synth_bench2022,synth_review2024,synth_longitudinal_review2025}.
Agent-centric evaluation demands something different: \emph{per-individual}
interpretability and \emph{query-answerable ground truth} for multi-hop
temporal questions.

\paragraph{Our approach: events as first-class objects.}
We introduce \dataset{}, an event-driven benchmark whose synthesis framework
treats temporal coherence and interpretability as first-class design goals.
The key idea: model each patient trajectory as a \emph{baseline health state
plus a superposition of discrete life and clinical events}.
Every event carries explicit temporal dynamics---sigmoid onset for gradual
initiation, exponential fade-out for recovery---that are both
human-interpretable and algorithmically verifiable.
Multiple events combine through superposition with saturation and projection
constraints, yielding a transparent decomposition: observation = baseline +
autoregressive residual + event contributions + noise.
Why is synthesis viable here? Because the benchmark evaluates \emph{temporal
reasoning} over structured event--indicator relationships, not clinical
fidelity. What matters is realistic causal structure and statistical
plausibility, both enforced by
construction~\cite{hernanrobins_whatif}.

\paragraph{How this enables a strong benchmark.}
From the event-driven representation we extract a ground-truth
event--indicator--time graph and derive 10{,}000 evaluation queries across
five user-centric dimensions---Lookup, Trend, Comparison, Anomaly, and
Explanation---each stratified into Easy, Medium, and Hard tiers.
This two-axis design isolates distinct failure modes (retrieval precision,
temporal alignment, arithmetic, evidence grounding) and discriminates among
retrieval and memory paradigms under controlled query distributions.
Easy queries? Most methods handle them competently. Medium and Hard queries
progressively expose the limits of memory-based retrieval---so benchmark
scores track genuine analytical capability, not prompt sensitivity.

\paragraph{Contributions.}
\dataset{} makes four contributions:
\begin{enumerate}
\item \textbf{A longitudinal synthetic benchmark with verifiable ground
truth.}
100 synthetic users, each spanning 1--5~years of daily device trajectories,
sparse exam visits, and structured life events.
Event--indicator relationships are defined by construction through explicit
temporal kernels, so ground truth is computable directly from the exported
structures (\cref{sec:benchmark}).

\item \textbf{A capability-discriminating evaluation design.}
Five query dimensions---Lookup, Trend, Comparison, Anomaly, and
Explanation---capture progressively harder temporal reasoning operations, each
stratified into three difficulty tiers for fine-grained diagnosis.
All questions derive deterministically from the event--indicator--time graph,
so difficulty reflects genuine reasoning demands rather than prompt
sensitivity (\cref{sec:benchmark}).

\item \textbf{An event-driven synthesis framework with explicit reliability
mechanisms.}
A hybrid pipeline delegates sparse semantic content (profiles, events, exam
narratives) to LLM modules while dense indicator trajectories are simulated
algorithmically under hard physiological constraints.
A trajectory planning step produces a multi-phase narrative arc that guides
event scheduling.
Four context engineering strategies---profile-conditioned population sampling,
multi-step decomposition, chain-of-thought clinical reasoning, and
human-calibrated marginal distribution validation---support the reliability of
LLM-generated
components (\cref{sec:framework,sec:framework:reliability}).

\item \textbf{Empirical evidence of capability stratification.}
DB agents (48--58\%) substantially outperform memory RAG baselines
(30--38\%), with the gap concentrated on Comparison and Explanation
queries requiring multi-hop reasoning and evidence
attribution (\cref{sec:experiments}).
\end{enumerate}

The remainder of the paper is organized as follows: \cref{sec:related} surveys
related work; \cref{sec:benchmark} defines the benchmark structure and
evaluation design; \cref{sec:framework} describes the event-driven synthesis
pipeline; \cref{sec:dataset-summary} reports dataset statistics;
\cref{sec:experiments} presents experiments and analysis; and
\cref{sec:conclusion} concludes.

\section{Related Work}
\label{sec:related}

\subsection{Synthetic longitudinal health data}

Three families of generators dominate synthetic health data.
Rule-based simulators, led by Synthea, offer transparent clinical logic and
controllable patient lifespans; they were not designed, however, for dense
event-aligned daily device signals~\cite{walonoski2018synthea}.
Deep generative models---medGAN, TimeGAN, EHR-M-GAN,
SynTEG---push distributional fidelity for specific modalities but keep their
causal mechanisms opaque, making event-aligned aggregation and
attribution-style evaluation
impractical~\cite{choi2017medgan,yoon2019timegan,ehrmgan2023,synteg2021}.
LLM-driven synthesis (e.g., LLMSYN) is flexible for narratives and structured
fields; guaranteeing long-horizon temporal coherence without explicit dynamics
remains an open challenge~\cite{hao2024llmsyn}.
Wearable-specific efforts have concentrated on IMU augmentation and activity
recognition---a different problem from longitudinal question answering.

Evaluation practices have grown more
sophisticated~\cite{synth_bench2022,synth_review2024,synth_longitudinal_review2025},
but the focus stays at the population level: marginal distributions and
downstream utility rather than individual-trajectory interpretability.
Privacy concerns add another
layer---membership inference attacks have been demonstrated against synthetic
health data~\cite{zhang2022mia,synteg2021}.
\dataset{} departs from this line by treating events as first-class objects
with explicit temporal kernels, producing an event--indicator--time graph that
supports controllable, auditable benchmark construction.

\subsection{Clinical benchmarks, longitudinal QA, and medical agents}

We are not aware of any benchmark that jointly covers a multi-year temporal
horizon, multi-source modality (device and clinical), and verifiable
attribution scoring.
MIMIC-IV and EHRSHOT underpin much of clinical ML but are single-modality and
lack event-level attribution ground
truth~\cite{mimiciv2022,ehrshot2023}.
emrQA and EHRSQL tackle question answering over medical records with temporal
constraints; both operate within short encounter windows and have no device
data~\cite{emrqa2018,lee2022ehrsql}.
Clinical TempEval and TIMER address temporal grounding---event ordering,
primarily---rather than quantitative indicator
attribution~\cite{bethard2016clinicaltempeval,timer2025}.

Medical agent benchmarks are proliferating:
MedAgentBench offers a FHIR-compliant virtual EHR~\cite{medagentbench2025},
MedJourney tests end-to-end clinical journeys~\cite{medjourney2024},
MedAgentBoard probes multi-agent collaboration~\cite{medagentboard2025}, and
AgentEHR targets EHR-native settings~\cite{agentehr2026}.
All test relevant skills. None, however, resolves the fundamental ambiguity of
attribution-style longitudinal questions in real EHR, where evidence is
incomplete and causal structure unknown.
\dataset{} complements them by targeting the setting where
(i)~device indicators and sparse exams coexist,
(ii)~plausibility is enforced by explicit constraints, and
(iii)~event--indicator relationships are defined by construction---enabling
scalable, difficulty-graded evaluation with deterministic ground truth.

\subsection{Retrieval, memory, and graph-based methods}

RAG and its memory- and graph-augmented variants pair LLMs with external
knowledge
stores~\cite{rag2020,graphrag2024,hipporag2024,lightrag2024,dygrag2025};
their differences become consequential in longitudinal health workloads.
Vanilla RAG indexes chunks by semantic similarity---effective for topical
retrieval, but prone to missing temporally distant passages that are causally
relevant~\cite{rag2020}.
HippoRAG maintains persistent, incrementally updated memory that can bridge
distant episodes; temporal alignment across heterogeneous sources is not
enforced~\cite{hipporag2024}.
Graph-based variants (GraphRAG, LightRAG, DyG-RAG) structure documents as
entity--relation graphs supporting multi-hop traversal, but their edges
encode semantic rather than temporal relations, leaving time-windowed
aggregation and cross-source joins
implicit~\cite{graphrag2024,lightrag2024,dygrag2025,graphrag_survey2025}.

Retrieval alone does not suffice for longitudinal health. Agents must align
time slices across sources, normalize units, join events with indicators and
visits, compute window statistics, and report auditable evidence---operations
that penalize flat similarity search, expose the limits of single-graph
traversal, and demand numerical rather than purely textual reasoning.
\dataset{} is designed to tease apart exactly these capabilities under one
scoring protocol.

\medskip
\noindent
Taken together, the picture has three gaps: synthetic generators lack
event-level causal mechanisms; clinical benchmarks do not jointly cover
multi-year, multi-source trajectories with verifiable attribution; and
retrieval, memory, and graph methods have not been systematically compared on
the temporal operations that longitudinal health demands.
\dataset{} bridges these gaps with a controllable generator whose events carry
first-class temporal semantics, a multi-source benchmark with programmatic
ground truth, and a dimension--tier taxonomy that isolates the operations
where current methods diverge.

\section{Benchmark: Structure and Evaluation Design}
\label{sec:benchmark}

\begin{figure}[t]
\centering
\includegraphics[width=0.8\textwidth]{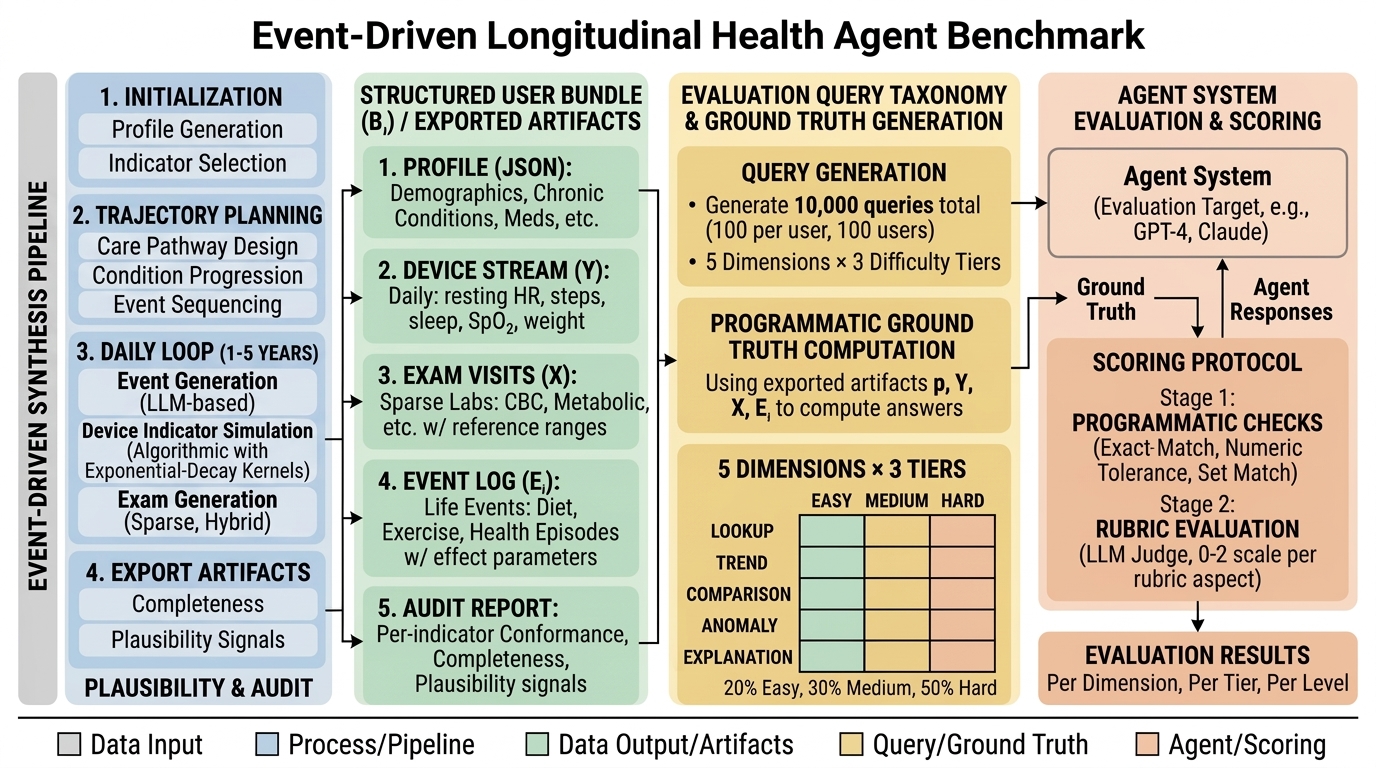}
\caption{Structure and evaluation design of the event-driven synthetic
longitudinal benchmark.}
\label{fig:benchmark-structure}
\end{figure}

We begin with what \dataset{} contains and what it measures, deferring the
generation process to \cref{sec:framework}.
The section covers the per-user data bundle, the event--indicator schema that
makes ground truth computable, the evaluation taxonomy (five dimensions,
three difficulty tiers), and the scoring
protocol (\cref{fig:benchmark-structure}).

\begin{figure}[t]
\centering
\includegraphics[width=\textwidth]{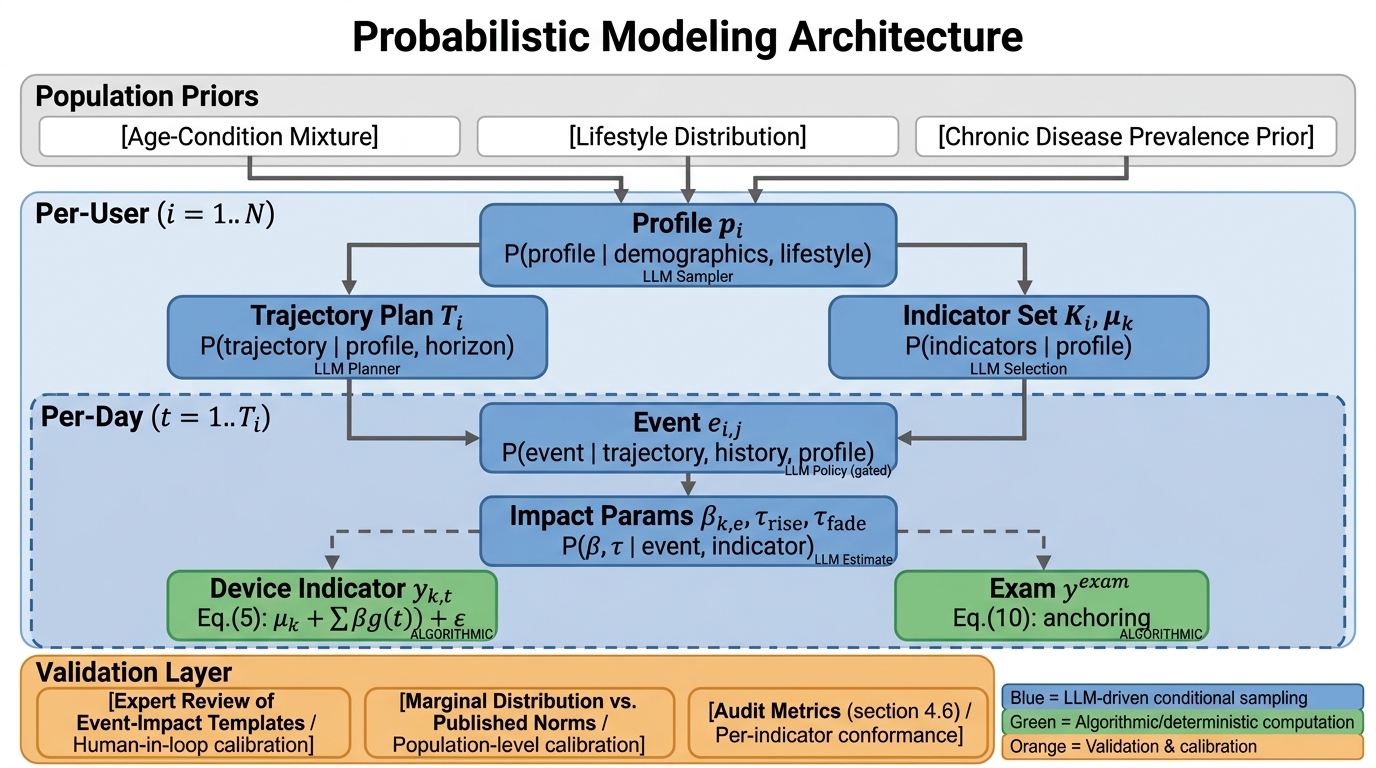}
\caption{Internal probabilistic modeling architecture.
Blue nodes denote LLM-driven conditional sampling; green nodes denote
algorithmic simulation governed
by \cref{eq:dynamics,eq:kernel,eq:exam_anchor}; orange nodes denote
validation mechanisms.
Each LLM call conditions on a narrow context (short history window, single
event type) to estimate a low-dimensional conditional distribution, while the
simulator computes dense daily dynamics deterministically.
The three-level validation layer covers expert review of event-impact
templates, population-level marginal calibration against published norms,
and per-indicator conformance auditing.}
\label{fig:prob-architecture}
\end{figure}

\subsection{User bundle structure}
\label{sec:benchmark:bundle}

For each user $i$ spanning $T_i$ days, \dataset{} exports a structured bundle
\begin{equation}
B_i = \big(p_i,\; \mathcal{T}_i,\; Y_i,\; X_i,\; E_i,\; audit_i\big),
\end{equation}
comprising six components:
\begin{itemize}
\item \textbf{Profile} $p_i$: demographics, chronic conditions, lifestyle
factors, and medication history, stored as structured JSON.
\item \textbf{Trajectory plan} $\mathcal{T}_i$: a narrative health arc
consisting of multiple temporal phases (approximately one per 90 days; see
\cref{sec:dataset-summary} for empirical statistics), each specifying a date
range and a description of the dominant health theme for that period (e.g.,
baseline stabilization, acute episode, recovery).
The trajectory plan guides event scheduling and ensures longitudinal narrative
coherence across phases.
\item \textbf{Device stream}
$Y_i=\{y_{i,k}(t)\}_{t=1,\dots,T_i,\; k\in K_i^{(d)}}$: daily values for
device indicators such as resting heart rate, step count, sleep duration,
SpO$_2$, and weight.
\item \textbf{Exam visits}
$X_i=\{x_{i,k}(t)\}_{t\in V_i,\; k\in K_i^{(e)}}$: sparse clinical
observations (CBC, metabolic panel, lipid panel, etc.)\ on visit days
$V_i\subset\{1,\dots,T_i\}$, each with reference ranges and normal/abnormal
status.
\item \textbf{Event log} $E_i=\{e_{i,j}\}_{j=1}^{J_i}$: life events (diet
changes, exercise routines, health episodes, long-term habits) with start
date, duration, affected indicator set, and per-indicator effect parameters.
\item \textbf{Audit report} $audit_i$: per-indicator, per-window conformance,
completeness, and plausibility signals that track generation quality.
\end{itemize}
All six components are exported as structured artifacts; queries and
ground-truth answers are computed directly from them, with no dependence on
external clinical knowledge.

\subsection{Event--indicator schema}
\label{sec:benchmark:schema}

\dataset{} rests on one design principle: \emph{events explicitly drive
indicator changes through deterministic temporal mechanisms}.
\cref{fig:prob-architecture} shows the architecture. LLM modules serve as
conditional samplers for low-dimensional decisions (profile, event, impact
parameters); an algorithmic simulator then computes dense daily dynamics under
explicit equations.
Each event $e$ specifies affected indicators, a signed magnitude
$\beta_{k,e}$ per indicator, and timing parameters controlling onset speed
and fade-out rate---together forming an impulse-response kernel whose effect
rises after the event starts and decays after it ends.

For example, the event ``started jogging routine'' (days 45--90) might affect
\{resting\_hr, steps, deep\_sleep\_ratio, active\_energy\}, with resting heart
rate decreasing ($\beta < 0$) and step count increasing ($\beta > 0$).
When multiple events overlap, their effects combine additively with a soft cap
that prevents implausible excursions.

A key consequence: because the event log records every event--indicator
relationship with explicit parameters, ground truth for any query---including
attribution---is programmatically computable.
We call this property \emph{mechanism recovery}: given an observed indicator
change, identify which generator-defined events contributed and rank them by
magnitude.
Mathematical details of the temporal kernel and simulation dynamics appear in
\cref{sec:framework}.

\subsection{Evaluation query taxonomy}
\label{sec:benchmark:queries}

Each user is paired with 100 evaluation queries (10{,}000 in total),
organized along two axes: five user-centric \emph{dimensions} and three
\emph{difficulty tiers}.
All queries derive deterministically from the event--indicator--time structure
of each user bundle.
\cref{tab:dimensions} lists the dimensions with their core operations;
\cref{tab:query_examples} gives representative examples.
A complete subtype inventory is available in the codebase.

\paragraph{Five dimensions.}
The dimensions mirror the kinds of questions that patients and clinicians
naturally ask about a longitudinal health trajectory:
\begin{itemize}
\item \textbf{Lookup} --- ``What is the value of X?''
Direct data retrieval over profile attributes, device values on a given date,
exam results, and event properties.
\item \textbf{Trend} --- ``How is X changing over time?''
Temporal pattern analysis including monthly aggregation, rate of change,
consecutive streaks, volatility, and regime detection.
\item \textbf{Comparison} --- ``How does A compare with B?''
Cross-event or cross-source comparisons such as pre-event vs.\ post-event
indicator deltas, shared indicator overlap between events, and relative
severity ranking.
\item \textbf{Anomaly} --- ``Is anything abnormal?''
Abnormality detection including threshold exceedance, abnormal streaks,
co-occurring abnormal indicators, and cross-exam deterioration tracking.
\item \textbf{Explanation} --- ``Why did this happen?''
Causal attribution and evidence organization: event contribution ranking,
counterfactual estimation, dominant event identification, and multi-event net
attribution.
\end{itemize}

\paragraph{Three difficulty tiers.}
Queries within each dimension are stratified into Easy, Medium, and Hard.
Tier boundaries are calibrated empirically: we evaluate a strong baseline
agent on a held-out generation session and assign each query generator to the
tier whose target accuracy band it falls into.
Easy queries test fundamental retrieval and single-step reasoning.
Medium queries require multi-step operations---cross-source joins, monthly
aggregation, multi-condition filtering.
Hard queries push further into temporal reasoning, mechanism-level
attribution, and multi-constraint chains.
Each user receives 20 queries per dimension; within each dimension the default
split is 20\% Easy, 30\% Medium, 50\% Hard, weighting the benchmark toward
tasks that most sharply discriminate architectures.

\paragraph{Dimension--tier interaction.}
Two axes give finer diagnostic resolution than a single difficulty ladder.
Two agents can match on aggregate accuracy yet diverge in their dimension
profiles---one may excel at Trend but fail on Explanation, another the
reverse.
Within a single dimension, the Easy-to-Hard gradient tells us whether
failures stem from insufficient retrieval (Easy) or limited reasoning
depth (Hard).

\begin{table*}[t]
\centering
\caption{Five evaluation dimensions stratified by three difficulty tiers.
Each cell summarizes the core operations tested at that dimension--tier
combination.}
\label{tab:dimensions}
\small
\begin{tabular}{l p{4.2cm} p{4.2cm} p{4.2cm}}
\toprule
Dimension & Easy & Medium & Hard \\
\midrule
Lookup
& Profile attribute, device value on date, exam indicator status
& Event-to-indicator mapping, yearly event/exam counts, indicator-to-event listing
& Most indicator-rich event, multi-constraint chain queries, most impacted indicator \\
\addlinespace
Trend
& Best/worst month, quarter means, seasonal pattern detection
& Monthly improvement, recovery detection, EWMA vs.\ SMA divergence
& Piecewise trend breakpoints, rolling z-score outlier count, regime change detection \\
\addlinespace
Comparison
& Two-event comparison, post-event change, relative change ratio
& Event severity ranking, indicator event-vs-baseline ratio, pre/post event comparison
& Shared indicators between events, symmetric difference, events affecting both indicators \\
\addlinespace
Anomaly
& Ever-abnormal check, abnormal streak, threshold exceedance days
& Abnormal trend acceleration, cross-exam persistence, deterioration rate
& Causal chain detection, abnormal periodicity, multi-indicator abnormal cluster \\
\addlinespace
Explanation
& Active events at anomaly, event contribution ranking on date, net indicator effect
& Impact ranking by magnitude, treatment effectiveness, recovery timeline
& Counterfactual attribution, dominant event identification, multi-event net attribution \\
\bottomrule
\end{tabular}
\end{table*}

\begin{table*}[t]
\centering
\caption{Representative query examples across dimensions and difficulty
tiers.}
\label{tab:query_examples}
\small
\begin{tabular}{l l p{10cm}}
\toprule
Dimension & Tier & Example Query \\
\midrule
Lookup & Easy & ``What was this user's resting heart rate on 2024-03-15?'' \\
Lookup & Med. & ``Which indicators were affected by the event \emph{started jogging routine}?'' \\
Lookup & Hard & ``Which event affected the most indicators?'' \\
\addlinespace
Trend & Easy & ``In which month was the average step count highest?'' \\
Trend & Med. & ``In which month did resting heart rate show the largest month-over-month change?'' \\
Trend & Hard & ``At which date did resting heart rate exhibit a regime change?'' \\
\addlinespace
Comp. & Easy & ``How did mean step count change after the event \emph{started jogging routine} compared with the 14 days before?'' \\
Comp. & Med. & ``What is the ratio of mean resting heart rate during the event \emph{high-sodium diet} to the baseline value?'' \\
Comp. & Hard & ``Which events share at least one affected indicator with the event \emph{high-sodium diet}?'' \\
\addlinespace
Anom. & Easy & ``Has fasting glucose ever been marked abnormal?'' \\
Anom. & Med. & ``For which exam indicator did the abnormal rate deteriorate most between the first and last exam?'' \\
Anom. & Hard & ``Which abnormal indicators co-occurred in a cluster across at least two consecutive exams?'' \\
\addlinespace
Expl. & Easy & ``Which events were active when resting heart rate exceeded two standard deviations above baseline?'' \\
Expl. & Med. & ``Rank events by their impact magnitude on resting heart rate during 2024-Q2.'' \\
Expl. & Hard & ``Rank all events by their estimated contribution to the observed drop in fasting glucose.'' \\
\bottomrule
\end{tabular}
\end{table*}

\subsection{Scoring protocol}
\label{sec:benchmark:scoring}

A single two-stage scoring protocol applies to all queries regardless of
dimension, combining programmatic checks with LLM-based rubric evaluation.

\paragraph{Stage 1: Programmatic checks.}
Each response is parsed into a canonical JSON schema (answer type, values,
dates, unit, source, optional evidence) and compared against ground truth.
Numeric answers allow tolerance-based matching to accommodate rounding:
\begin{equation}
|\hat{v}-v|\le \max(\epsilon_{\mathrm{abs}},
\epsilon_{\mathrm{rel}}\cdot |v|),
\end{equation}
with $\epsilon_{\mathrm{abs}}=0.01$ and
$\epsilon_{\mathrm{rel}}=0.01$.
For set-valued answers, we check exact set match.
A response that fails any programmatic check receives a score of zero.

\paragraph{Stage 2: Rubric evaluation.}
Responses that pass Stage~1 are scored by an LLM judge (GPT-4.1) under fixed seeds.
Pure programmatic matching is insufficient here: agents express correct
answers in varied list orderings, alternative numerical representations, and
different phrasing conventions.
The rubric assigns a 0--2 quality score whose aspects vary by dimension.
Lookup and Anomaly queries are judged on value correctness and format
compliance; Trend and Comparison add statistical reasoning and comparison
logic; Explanation queries are additionally evaluated for \emph{baseline
clarity}, \emph{evidence ordering rationale}, and \emph{non-causal language}.
The final per-query score is the product of the binary programmatic gate and
the mean rubric score normalized to $[0,1]$.
Rubric definitions and the judge prompt are summarized in \cref{app:prompts};
full versions ship with the codebase.

\subsection{Design rationale}
\label{sec:benchmark:rationale}

The design targets the three evaluation gaps identified
in \cref{sec:related}.
Synthetic generation sidesteps the privacy and distribution barriers of real
cohorts, making evaluation open and reproducible.
The explicit event--indicator schema with temporal kernels makes attribution
ground truth computable by construction---a property that real observational
data, with its ambiguous event boundaries and unmeasured confounders, cannot
offer.
The dimension--tier taxonomy then isolates the capabilities that existing
benchmarks under-test---cross-source comparison, temporal trend analysis,
abnormality detection, and evidence-structured attribution---at graded
difficulty levels that reveal where architectures diverge.

\section{Event-Driven Synthesis Framework}
\label{sec:framework}

With the benchmark specification in place (\cref{sec:benchmark}), we turn to how the data are generated.
The pipeline is hybrid: LLM modules handle sparse semantic content---profiles, event narratives, exam metadata---while an algorithmic simulator produces dense daily device trajectories under hard constraints.

\subsection{Overview}
\label{sec:framework:overview}

\begin{figure}[t]
\centering
\includegraphics[width=0.8\textwidth]{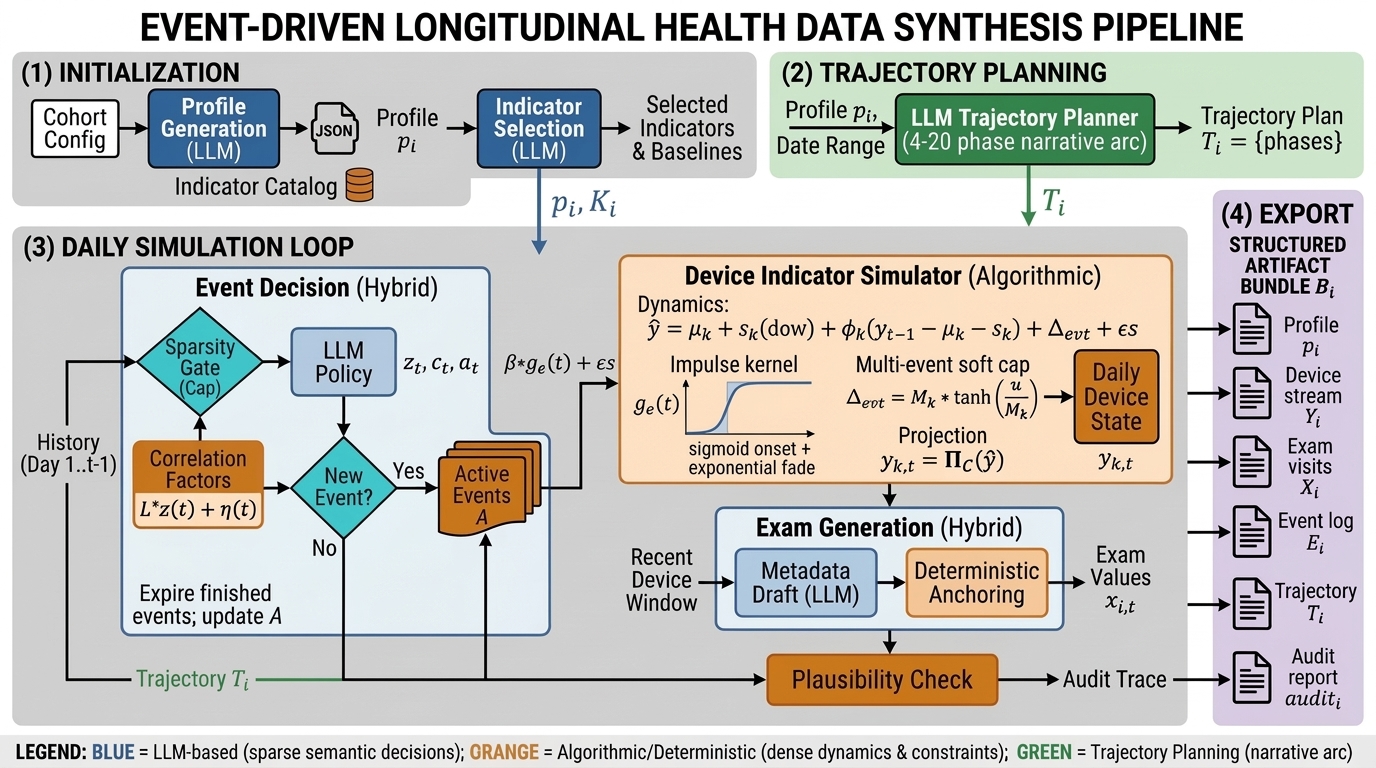}

\caption{Hybrid generation pipeline. LLM modules handle sparse semantic
decisions (profiles, trajectory plan, event narratives, exam metadata),
while algorithmic simulation produces daily device indicators under explicit
dynamics and deterministic constraints.
(1)~Initialization with Profile Generation and Indicator Selection;
(2)~Trajectory Planning that produces a multi-phase narrative arc;
(3)~Daily Loop with Event Decision (LLM + trajectory context + sparsity gate),
Device Indicator Simulator (algorithmic), and Exam Generation (LLM +
deterministic anchoring);
(4)~Export producing structured artifacts.}
\label{fig:pipeline}
\end{figure}

Each user bundle $B_i$ (\cref{sec:benchmark:bundle}) is produced in two
stages: a one-time trajectory planning step that lays out the longitudinal
narrative arc, then a daily simulation loop.
Within each simulated day, three modules fire in sequence: event generation
(LLM-based, conditioned on the trajectory plan), device indicator simulation
(algorithmic), and exam generation (hybrid).
\cref{fig:pipeline} shows the pipeline;
\cref{fig:prob-architecture} the probabilistic dependencies;
\cref{alg:synthesis} the per-user generation loop.

\begin{algorithm}[t]
\caption{\dataset{} synthesis routine (per user).}
\label{alg:synthesis}
\begin{algorithmic}[1]
\Require horizon $T_i$, sparsity level $s$ (weekly cap $M(s)$), indicator sets $K^{(d)},K^{(e)}$
\Ensure bundle $B_i=(p,\mathcal{T},Y,X,E,audit)$
\State Sample profile $p$; initialize baselines $\{\mu_k\}$ and initial device state $\{y_{k,0}\}$
\State $\mathcal{T} \gets \mathrm{LLM\_trajectory\_plan}(p, T_i)$ \Comment{multi-phase narrative arc (mean 11, range 4--20)}
\State Initialize active events $A\gets\varnothing$, event log $E\gets\varnothing$, device log $Y\gets\varnothing$, exam log $X\gets\varnothing$, audits $audit$
\For{$t=1$ to $T_i$}
  \Statex \hspace{\algorithmicindent}\textit{\% (A) Event generation (capped)}
  \If{weekly-start count $< M(s)$}
    \State $(z, \mathrm{mark}) \gets \mathrm{LLM\_policy}(p, \mathcal{T}, \text{history up to day } t)$
    \If{$z=1$} instantiate event $e$ from mark; add $e$ to $A$; append $e$ to $E$ \EndIf
  \EndIf
  \Statex \hspace{\algorithmicindent}\textit{\% (B) Device indicators generation}
  \For{each device indicator $k\in K^{(d)}$}
    \State Compute event input from active events via kernel $g_e(t)$
    \State Draw correlated day-level noise $\epsilon_{k,t}$ (shared global/group factors + idiosyncratic)
    \State Update $y_{k,t}$ using \cref{eq:dynamics} (with overlap handling \cref{eq:softcap})
  \EndFor
  \State Append day $t$ device values to $Y$; update audits (violations, clipping, coverage)
  \Statex \hspace{\algorithmicindent}\textit{\% (C) Exam generation (sparse)}
  \If{exam scheduled on day $t$}
    \State LLM drafts structured metadata + discrete interpretations conditioned on active events
    \State Anchor numeric fields to recent device windows (\cref{eq:exam_anchor}); derive ranges/status deterministically
    \State Append exam to $X$; update exam--device consistency audits
  \EndIf
  \State Expire events whose fade-out window ended; update $A$
\EndFor
\State \Return $(p, \mathcal{T}, Y, X, E, audit)$
\end{algorithmic}
\end{algorithm}

\subsection{Trajectory planning}
\label{sec:framework:trajectory}

Before the daily loop begins, a trajectory plan $\mathcal{T}_i$ is generated
to provide a coarse narrative arc spanning the full observation period.
Given the user profile $p_i$, the date range $(t_1, t_{T_i})$, and any
pre-existing initial events, the LLM outputs an overall narrative (3--5
sentences on the health arc and key turning points) together with a sequence
of temporal phases---e.g., ``baseline stabilization with gradual weight
management,'' ``acute respiratory episode and recovery,'' ``sustained exercise
routine with cardiovascular improvement.''
Phase count scales with the observation horizon, roughly one per 90 days.

The plan feeds into the downstream pipeline in two ways.
It gives the event decision agent (\cref{sec:framework:events}) phase-level
context---the agent conditions on the current phase description when deciding
whether to spawn a new event, steering types and timing toward narrative
coherence.
It also produces an auditable record of the intended storyline, so domain
experts can verify medical plausibility at the arc level before drilling into
individual events or indicator traces.

\subsection{Event generation}
\label{sec:framework:events}

Two mechanisms jointly constrain event generation: the trajectory plan
$\mathcal{T}_i$ (\cref{sec:framework:trajectory}) and a hard sparsity gate.
The gate enforces a rolling weekly cap $M(s)$---if the trailing 7-day window
already contains $M(s)$ event starts, the day is skipped regardless of
context.
The LLM policy runs only when the gate passes.

Formally, let $u=\phi(p)\in\mathbb{R}^D$ be an embedding of the user profile
$p$, let $\mathcal{H}_t$ denote the generation history up to day $t$ (active
events, recent device values, most recent exam, and calendar features), and
let $\tau_t$ denote the trajectory phase that covers day $t$.
We model event occurrence in discrete time with a gated Bernoulli process plus
a categorical mark:
\begin{align}
z_t &\sim \mathrm{Bernoulli}\!\bigl(g_t \cdot
  \sigma\!\bigl(w_0^\top \psi(u,\tau_t,\mathcal{H}_t)\bigr)\bigr),\\
c_t \mid z_t\!=\!1 &\sim \mathrm{Categorical}\!\bigl(
  \mathrm{softmax}\bigl(W \psi(u,\tau_t,\mathcal{H}_t)\bigr)\bigr),
\end{align}
where $g_t\in\{0,1\}$ is the deterministic sparsity gate (encoding the weekly
cap, maximum active events, and warm-up constraints), and $\tau_t$ injects the
current phase description into the feature map $\psi$.
Conditional on $(z_t,c_t)$, the LLM policy generates the mark attributes
$a_t$---duration, affected indicators, narrative, signed per-indicator
magnitudes $\beta_{k,e}$, and kernel timing parameters---subject to schema
constraints.

The trajectory plan steers events through a three-tier priority scheme.
\emph{Storyline events} directly realize the current phase theme---e.g., a
medication adjustment during ``treatment escalation''---and receive highest
priority.
\emph{Texture events} (seasonal colds, social dinners, minor injuries) are
welcomed so long as they do not contradict the trajectory direction; they add
realism and diversity.
The policy also watches for \emph{trajectory gaps}: if a phase describes a
change (e.g., ``post-injury recovery'') that no recent event has realized, gap
filling is prioritized.
Together, these rules keep the realized event sequence aligned with the
narrative arc while preserving the variety of a realistic timeline.

Events remain active during their duration and continue to influence
indicators through a fade-out window---a property that directly shapes the
temporal patterns probed by windowed queries.

\subsection{Device indicators generation}
\label{sec:framework:dynamics}

Each daily device indicator is modeled as a constrained stochastic dynamical
system driven by event inputs.
Let $\mu_k$ denote the personalized baseline for indicator $k$ and
$s_k(\dow_t)$ a weekday/seasonality term.
The unconstrained proposal on day $t$ is
\begin{equation}
\hat{y}_{k,t}
\;=\;
  \mu_k + s_k(\dow_t)
  + \phi_k\big(y_{k,t-1}-\mu_k-s_k(\dow_{t-1})\big)
  + \Delta^{(\mathrm{evt})}_{k,t}
  + \epsilon_{k,t},
\label{eq:dynamics}
\end{equation}
where $\phi_k\in[0,1)$ controls inertia and mean reversion around the
baseline residual, $\Delta^{(\mathrm{evt})}_{k,t}$ aggregates event effects
(defined below), and $\epsilon_{k,t}$ captures day-level noise.

Hard plausibility is enforced by projecting values into a feasible set that
jointly constrains range and day-to-day slope:
\begin{equation}
y_{k,t} = \Pi_{\mathcal{C}_k}\!\big(\hat{y}_{k,t}\big),\quad
\mathcal{C}_k = \{y:\ L_k \le y \le U_k,\;
|y - y_{k,t-1}| \le \Delta_k\},
\label{eq:projection}
\end{equation}
where $[L_k, U_k]$ is the physiological range and $\Delta_k$ is a per-indicator
slope limit that prevents unrealistic day-to-day jumps.
The audit report records violation statistics on $\hat{y}_{k,t}$ before
projection.

\paragraph{Impulse-response kernel.}
Each event $e$ with start day $t_{\mathrm{start},e}$, end day
$t_{\mathrm{end},e}=t_{\mathrm{start},e}+d_e$, onset time $\tau_{\mathrm{rise},e}$,
and fade-out window $\tau_{\mathrm{fade},e}$ contributes through a piecewise
kernel:
\begin{equation}
g_e(t) =
\begin{cases}
0, & t \le t_{\mathrm{start},e},\\[3pt]
\sigma\!\bigl(k_e\,((t-t_{\mathrm{start},e})-\tfrac{\tau_{\mathrm{rise},e}}{2})\bigr),
  & t_{\mathrm{start},e} < t \le t_{\mathrm{end},e},\\[3pt]
\exp\!\bigl(-\alpha_e\,(t-t_{\mathrm{end},e})\bigr),
  & t_{\mathrm{end},e} < t \le t_{\mathrm{end},e} + \tau_{\mathrm{fade},e},\\[3pt]
0, & t > t_{\mathrm{end},e} + \tau_{\mathrm{fade},e},
\end{cases}
\label{eq:kernel}
\end{equation}
with steepness $k_e = 6/\tau_{\mathrm{rise},e}$ and decay rate
$\alpha_e = 3/\tau_{\mathrm{fade},e}$.
The sigmoid phase models gradual onset; the exponential phase models
post-event recovery or waning.

\paragraph{Multi-event superposition and saturation.}
When multiple events overlap, na\"ive linear stacking can push indicators
into implausible excursions even before projection.
A smooth soft-cap on the raw event sum addresses this:
\begin{equation}
u_{k,t}=\sum_{e\in\mathcal{E}_t}\beta_{k,e}\,g_e(t),
\qquad
\Delta^{(\mathrm{evt})}_{k,t}=M_k\,\tanh\!\left(\frac{u_{k,t}}{M_k}\right),
\label{eq:softcap}
\end{equation}
where $\mathcal{E}_t$ is the set of events active on day $t$ (including
fade-out) and $M_k$ controls the maximum plausible deviation from overlapping
events.
$\Delta^{(\mathrm{evt})}_{k,t}$ replaces $u_{k,t}$
inside \cref{eq:dynamics}, preserving additivity for small overlaps while
preventing runaway stacking.

\paragraph{Correlated day-level noise.}
To produce coherent day-to-day co-movement across related indicators, we draw
noise from a lightweight factor model:
\begin{equation}
\bm{\epsilon}(t) = \mathbf{L}\,\mathbf{z}(t) + \bm{\eta}(t),\quad
\mathbf{z}(t)\sim\mathcal{N}(\mathbf{0},\mathbf{I}),\;
\bm{\eta}(t)\sim\mathcal{N}(\mathbf{0},\mathbf{D}),
\label{eq:noise}
\end{equation}
where $\mathbf{L}$ is a low-rank loading matrix that captures shared global
and group-level factors, and $\mathbf{D}$ is a diagonal matrix of
idiosyncratic variances.
This ensures that related indicator groups (e.g., cardiovascular indicators)
co-move on the same day.

Hard min/max clipping often produces boundary jitter for bounded or non-negative indicators.
We therefore compute the update in a transformed coordinate system---identity, log, or logit, chosen per indicator type---then invert the transform and apply the projection $\Pi_{[L_k,U_k]}(\cdot)$ in \cref{eq:dynamics}.
This reduces boundary artifacts without sacrificing determinism or auditability.

\subsection{Exam indicators generation}
\label{sec:framework:exam}

Exams are sparse but not independent snapshots---they reflect the same
event-driven trajectory as the device stream.
Generation is hybrid. An LLM drafts structured metadata and discrete
interpretations (notable panels, abnormal findings, short clinical
impressions) conditioned on profile, recent device trends, and active events.
Numeric values are then anchored deterministically to stay consistent with the
device stream.

For an exam on day $t$ and indicator $k$, we compute a recent-window device statistic $\tilde{y}_k(t)$. The window length is short for fast indicators and longer for slow indicators such as metabolic and weight.
If the window has insufficient points, we fall back to a local latent truth derived from the baseline and the event drive at day $t$ (so events still influence the exam value even under sparse device coverage).
We then anchor:
\begin{equation}
y^{(\mathrm{exam})}_k(t)
\;=\;
\Pi_{[L_k,U_k]}\big(\tilde{y}_k(t)+\xi_k\big),
\label{eq:exam_anchor}
\end{equation}
where $\xi_k$ is a small deterministic perturbation seeded by $(\text{user},t,k)$.
Reference ranges and normal/abnormal status are derived deterministically from $\big(y^{(\mathrm{exam})}_k(t),\;\text{reference range}\big)$, and the LLM narrative is constrained to remain consistent with these derived results.

\subsection{Plausibility and audit artifacts}
\label{sec:plausibility}
\label{sec:constraints}

Plausibility is enforced through deterministic simulation constraints; a
per-user audit report tracks generation quality over time.
Audits follow the standard data-quality taxonomy of conformance, completeness,
and plausibility~\cite{kahn2016dq}.
Conformance ensures canonical indicator keys and consistent
units---UCUM~\cite{hl7ucum} keeps quantities machine-unambiguous.
Completeness tracks coverage and missingness, attaching data-absent-reason
codes rather than silently dropping values~\cite{fhirDataAbsentReason}.
Plausibility covers value validity (hard bounds), stability (projection
activation rate as a proxy for poor parameterization), and cross-source
consistency (device--exam agreement under the anchoring windows of
\cref{eq:exam_anchor}).
Failures are localized to specific indicators and time windows, making the
generator easy to tune without changing the benchmark interface.

\subsection{Reliability of LLM-driven synthesis}
\label{sec:framework:reliability}

LLMs drive three generative steps: profile creation, event decision and
impact estimation, and exam narrative drafting.
We do not claim that LLM-generated distributions perfectly mirror real-world
clinical populations. The claim is narrower: synthesis is \emph{reliable
enough for benchmarking purposes}. Below we describe the mechanisms behind
this claim.

Modern LLMs absorb vast medical and health corpora---clinical guidelines,
epidemiological studies, wearable-device research, patient
narratives---encoding implicit distributional knowledge about indicators,
event--indicator relationships, and population-level
variation~\cite{synth_review2024,synth_longitudinal_review2025}.
Generating a profile conditioned on demographic and lifestyle attributes
amounts to sampling from $P(\text{profile} \mid \text{demographics,
lifestyle})$ learned across this corpus; varying the conditioning variables
across the cohort acts as stratified sampling over population subgroups.

Four context engineering strategies sharpen this implicit knowledge
(\cref{fig:prob-architecture}):

\paragraph{Profile-conditioned population sampling.}
Each profile specifies demographics, chronic conditions, lifestyle factors,
and personality traits before any event or indicator is generated
(\cref{sec:benchmark:bundle}).
Three age strata, multiple chronic-disease combinations, and diverse
lifestyles push the generator into different regions of the population
distribution---analogous to stratified sampling in epidemiological
surveys---so that between-user variation reflects real demographic diversity
rather than a single mode of the generative model.

\paragraph{Multi-step decomposition into low-dimensional conditionals.}
Rather than generating a multi-year trajectory in one pass, generation is
decomposed into narrow conditional decisions.
Each event decision sees only a short history window (7-day device values,
active events, last exam); each impact estimation conditions on a single event
type and a small indicator set (\cref{sec:framework:events}).
The effective dimensionality of each LLM call is bounded: instead of modeling
the joint over thousands of indicator-days, the model estimates localized
probabilities $P(\text{impact} \mid \text{event type, indicator, profile})$
where medical knowledge is well-established---exercise--heart-rate
effects~\cite{reimers2018exercise_rhr}, sodium--blood-pressure
relationships~\cite{lai2022sodium_reduction_bp}, sleep--HRV
dynamics~\cite{zhang2025sleepdeprivation_hrv}.

\paragraph{Chain-of-thought reasoning for physiological plausibility.}
Every LLM generation step externalizes its clinical rationale before emitting
structured output.
For event decisions, the model reasons about whether the recent history and
active events make a candidate event physiologically likely, then commits to
the binary decision and timing parameters.
For impact estimation, it states the expected direction and approximate
magnitude of each affected indicator---citing the mechanism (e.g., ``acute
alcohol intake suppresses HRV via sympathetic activation'')---before outputting
$\beta$ and $\tau$ values.
Forcing explicit reasoning before commitment steers the LLM toward medically
grounded outputs and produces an auditable trace.
Empirically, omitting chain-of-thought leads to more frequent sign errors
(e.g., HRV increasing after sleep deprivation) and less consistent temporal
dynamics across runs.

\paragraph{Human-calibrated marginal distribution validation.}
Individual conditional probabilities may contain errors, but
\emph{aggregate} distributional properties can be checked against known
population statistics.
Calibration proceeds on two fronts.
Clinicians reviewed a representative sample of event--indicator impact
templates, verifying that direction, rough magnitude, and temporal dynamics
are medically plausible (\cref{app:prompts}).
Separately, marginal distribution auditing compares event frequencies,
indicator baseline ranges, and exam result distributions against published
reference values~\cite{fraser1989biologicalvariation,sandberg2024aps_bv}.
The human-approval rate on a random sample of generated relationships provides
empirical evidence of distributional plausibility without requiring
point-level accuracy.

\paragraph{Reliability scope.}
This reliability argument targets \emph{benchmarking validity}, not clinical
fidelity.
Ground truth derives from the generator-defined event--indicator schema, not
from claims about real-world effect sizes.
A method that fails to recover mechanism relationships under these controlled
dynamics is unlikely to succeed on noisier real data---a necessary-condition
argument analogous to unit testing in software engineering.
The strategies above ensure that the controlled dynamics themselves are
medically plausible, keeping benchmark performance informative about agent
capabilities in realistic scenarios.

\section{Dataset Statistics}
\label{sec:dataset-summary}

All statistics below are computed directly from the exported
artifacts---profiles, device records, exam visits, events, and audit
reports---and reported as per-user values aggregated across the cohort unless
stated otherwise.

\subsection{Cohort composition}
Demographics and chronic-condition prevalence appear in
\cref{tab:cohort_stats};
\cref{tab:age_condition_mix} lists the target mixture used during profile
sampling.
Three age strata cover the chronic-disease spectrum most relevant to consumer
wearables---emerging metabolic and mental-health risks in younger adults,
established cardiometabolic conditions in middle age, and multi-morbidity in
older adults---ensuring the event overlap and cross-indicator diversity needed
for Comparison and Explanation queries (\cref{sec:benchmark:queries}).

\begin{table}[t]
\centering
\caption{Cohort summary statistics.}
\label{tab:cohort_stats}
\begin{tabular}{l r}
\toprule
Metric & Value \\
\midrule
Total users & 100 \\
Age mean (std) & 43 (16) \\
Age range & 18--80+ \\
Male & 53\% \\
Female & 47\% \\
Users with $\ge 1$ chronic condition & 98\% \\
\bottomrule
\end{tabular}
\end{table}

\begin{table}[t]
\centering
\caption{Age group and condition distribution (target mixture).}
\label{tab:age_condition_mix}
\begin{tabular}{l l r}
\toprule
Age Group & Primary Conditions & Users \\
\midrule
18--44 & Obesity/Metabolic, Pre-diabetes, Depression/Sleep, Healthy & 33 \\
45--64 & T2DM, Hypertension, Metabolic, Depression, Comorbid & 44 \\
65--80+ & HTN+CVD, T2DM+CKD, Osteoarthritis & 23 \\
\bottomrule
\end{tabular}
\end{table}

\subsection{Trajectory plan statistics}
Each user bundle includes a trajectory plan $\mathcal{T}_i$
(\cref{sec:framework:trajectory}) consisting of on average 11 phases
(range 4--20), with a mean phase duration of 106 days.
The number of phases scales with the observation horizon: users spanning
roughly one year have 4--6 phases, while those spanning three or more years
have 12--17 phases.

\subsection{Longitudinal coverage}
Each user $i$ spans $T_i$ days on a daily grid;
\cref{tab:coverage_stats} reports trajectory length, total observation days,
exam density, device-day coverage (fraction of days with at least one numeric
observation), and indicator-level numeric coverage (fraction of $(t,k)$ pairs
with numeric values).
The 90.0\% numeric coverage is intentional: the generator attaches
data-absent-reason codes to missing values rather than silently dropping
them (\cref{sec:plausibility}).

\begin{table}[t]
\centering
\caption{Longitudinal coverage statistics.}
\label{tab:coverage_stats}
\begin{tabular}{l r}
\toprule
Metric & Value \\
\midrule
Time span per user (mean) & 1{,}196 days \\
Time span per user (median) & 1{,}189 days \\
Time span per user (range) & 388--1{,}813 days \\
Total observation days $\sum_i T_i$ & 119{,}600 \\
Exam visits per user (mean) & 7 \\
Exam visit density (mean) & 2 visits/year \\
Device-day coverage (mean) & 100\% \\
Indicator numeric coverage (mean) & 90.0\% \\
\bottomrule
\end{tabular}
\end{table}

\subsection{Event statistics}
Events drive the temporal and attribution queries that form the core of the
benchmark.
\cref{tab:event_stats} reports event counts, durations, and overlap.
The mean and median durations differ substantially (88 vs.\ 4 days) because
long-term habits and sustained exercise routines pull the mean
upward, while the majority of events are acute health episodes such as
tension headaches, mild gastroenteritis, and situational anxiety (median 2
days) or short-lived diet changes such as occasional late-night meals and
social dinners (median 21 days).
On average, 2--3 short-term events ($\le$90 days) are concurrently active on
any given day; including long-running habits and persistent lifestyle changes,
the total rises to about 9.
This high concurrency is by design: it ensures that attribution queries
(\cref{sec:benchmark:queries}) must disentangle multiple overlapping effects
rather than attribute changes to a single event.

\begin{table}[t]
\centering
\caption{Event statistics.}
\label{tab:event_stats}
\begin{tabular}{l r}
\toprule
Metric & Value \\
\midrule
Total events & 18{,}800+ \\
Events per user (mean) & 188 \\
Event categories & 4 \\
Mean duration (days) & 88 \\
Median duration (days) & 4 \\
Concurrent events per day (mean) & 2--3 short-term; $\sim$9 total \\
\bottomrule
\end{tabular}
\end{table}

\subsection{Indicator coverage}
Indicator breadth is summarized in \cref{tab:indicator_coverage}.
Device indicators fall into six physiological groups---sleep, cardiovascular,
metabolic, activity, weight, blood oxygen---with 3--8 indicators per group
depending on the condition profile.
Exam indicators cover standard laboratory panels (CBC, metabolic, lipid,
liver/renal, inflammatory).
Fifteen indicators are measured by both sources (blood pressure, glucose,
SpO$_2$, among others); these are anchored during exam generation
(\cref{sec:framework:exam}) and appear frequently in cross-source queries.

\begin{table}[t]
\centering
\caption{Indicator coverage.}
\label{tab:indicator_coverage}
\begin{tabular}{l r l}
\toprule
Type & Count & Examples / Frequency \\
\midrule
Device indicators & 159 (60/user) & Daily (sleep, cardio, metabolic, activity, weight, SpO$_2$) \\
Exam indicators & 322 (200/user) & Per visit (CBC, metabolic, lipid, liver, renal, inflammatory) \\
Cross-source overlap & 15 & BP, glucose, SpO$_2$, body composition, pulmonary \\
\bottomrule
\end{tabular}
\end{table}

\subsection{Audit metrics}
\begin{figure}[!htbp]
\centering
\includegraphics[width=0.95\textwidth]{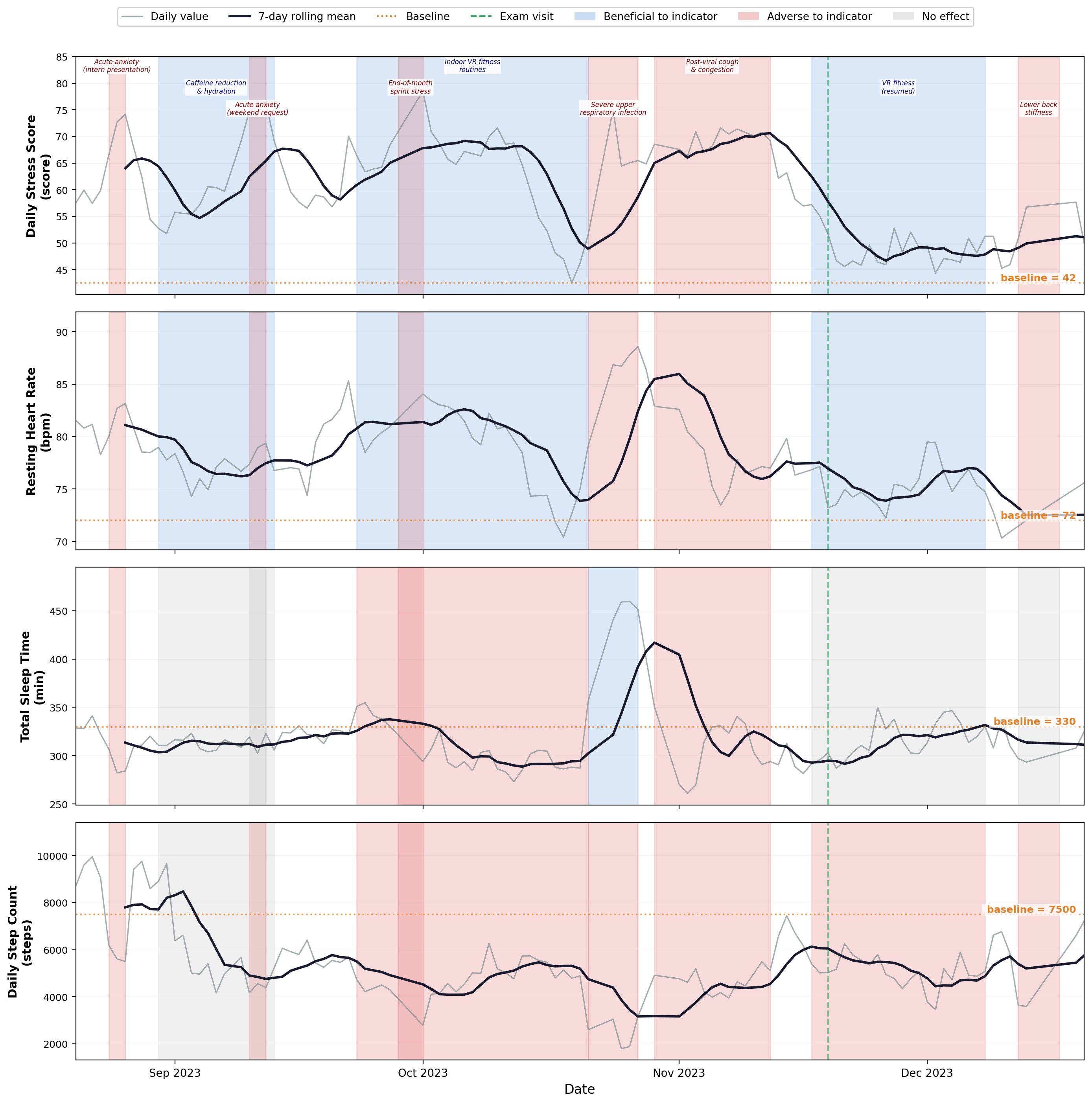}
\caption{Four-month trajectory excerpt for one synthetic user showing four device indicators (Daily Stress Score, Resting Heart Rate, Total Sleep Time, Daily Step Count) with nine labeled life events. Shaded regions mark active event periods; blue indicates a beneficial effect on that indicator, red an adverse effect, and gray no effect---so the same event may appear in different colors across panels (e.g., ``Indoor VR fitness routines'' is blue for Stress/HR but red for Sleep/Steps). The black line is the 7-day rolling mean; the orange dotted line marks each indicator's personalized baseline; the green dashed line marks an exam visit. Only the most prominent events are labeled; additional short-term events (e.g., acute anxiety episodes, OTC medication use) also contribute to the observed fluctuations.}
\label{fig:trajectory-example}
\end{figure}

Cohort-level audit metrics appear in \cref{tab:plausibility_metrics}.
Range and slope violation rates are measured on the unconstrained proposal
$\hat{y}_k(t)$ before projection (\cref{eq:dynamics}); clipping rate records
how often projection activates; exam--device consistency is checked on
overlapping indicators via the anchoring logic of \cref{sec:framework:exam}.
Every violation rate is zero; every consistency metric reaches 100\%.
This is by design. The soft-cap (\cref{eq:softcap}) and transform-domain
updates (\cref{sec:framework:dynamics}) keep proposals within plausible
bounds, so the hard projection $\Pi_{\mathcal{C}_k}$ rarely fires.
Perfect scores confirm that the constraint pipeline works as intended---not
that bounds are too loose; pre-projection violation counters would flag
parameterization problems if they existed.
\cref{fig:trajectory-example} illustrates a representative trajectory where the sigmoid onset and exponential fade-out of event effects are visible in the device stream, with exam observations anchored to the same underlying dynamics.

\begin{table}[t]
\centering
\caption{Aggregated audit metrics (means across users).}
\label{tab:plausibility_metrics}
\begin{tabular}{l r}
\toprule
Metric & Mean \\
\midrule
Indicator key presence rate & 100\% \\
Unit (UCUM) presence rate & 100\% \\
Range violation rate (pre-projection) & 0\% \\
Slope violation rate (pre-projection) & 0\% \\
Clipping rate (post-projection) & 0\% \\
Exam--device consistency (overlap indicators) & 100\% \\
\bottomrule
\end{tabular}
\end{table}

\section{Experiments}
\label{sec:experiments}

All experiments use the 10{,}000-query benchmark defined in
\cref{sec:benchmark} with its dimension--tier taxonomy and two-stage scoring
protocol.

\subsection{Baselines and input representations}

The comparison spans three paradigms (\cref{tab:baselines}): \emph{LLM
w/ tools} where each model calls a shared tool interface over user artifacts,
\emph{DB agents} that issue structured API calls (filter, aggregate, join)
over a DuckDB store, and \emph{Memory RAG} methods that augment retrieval
with memory architectures.
GPT-5.4 serves as the base LLM for all DB agent and Memory RAG methods;
LLM w/ tools entries test six models (GPT-5.2, GPT-5.4, Gemini 3 Flash
Preview, Sonnet 4.6, MiniMax M2.5, GLM-5) with the same tool interface---each
model receives the same tool suite (lookup, query, read, search) to avoid
context overflow from full artifact serialization.
Outputs follow a unified JSON schema for programmatic scoring.

\begin{table*}[t]
\centering
\caption{Baseline methods across three paradigms.}
\label{tab:baselines}
\small
\begin{tabular}{l l l p{7.5cm}}
\toprule
Paradigm & Method & Base LLM & Description \\
\midrule
\multirow{6}{*}{LLM w/ tools}
& GPT-5.2          & --    & \multirow{6}{7.5cm}{Each model uses its native tool-use capability with the same tool suite (lookup, query, read, search) over user artifacts. No agent framework; the LLM decides which tools to call.} \\
& GPT-5.4          & --    & \\
& Gemini 3 Flash   & --    & \\
& Sonnet 4.6       & --    & \\
& MiniMax M2.5     & --    & \\
& GLM-5            & --    & \\
\midrule
\multirow{3}{*}{DB agent}
& Theta General      & GPT-5.4 & \multirow{3}{7.5cm}{DB-native agent issuing structured API calls (filter, aggregate, join) over a DuckDB store of user artifacts. Three prompt variants with increasing domain specificity.} \\
& Theta Expert       & GPT-5.4 & \\
& Theta Smart Expert & GPT-5.4 & \\
\midrule
\multirow{4}{*}{Memory RAG}
& HippoRAG ($k{=}10$) & GPT-5.4 & \multirow{3}{7.5cm}{Hippocampus-inspired memory architecture with knowledge graph consolidation~\cite{hipporag2024}. Retrieval budget $k$ varied.} \\
& HippoRAG ($k{=}20$) & GPT-5.4 & \\
& HippoRAG ($k{=}50$) & GPT-5.4 & \\
\bottomrule
\end{tabular}
\end{table*}

\subsection{Main results}

Accuracy by dimension appears in \cref{tab:main_results_dim} and by
difficulty tier in \cref{tab:main_results_tier}.

\paragraph{LLM w/ tools: model capability matters.}
LLM w/ tools accuracy spans a wide range (45--63\%) depending on the underlying
model.
Gemini 3 Flash Preview leads at 62.9\%, with particularly strong Trend
performance (94.8\%); GPT-5.2 trails at 45.4\%.
All LLM w/ tools share the same tool interface, so the gap is attributable to
model-level reasoning and tool-use proficiency rather than retrieval design.

\paragraph{DB agents: strong on structured queries, mixed on temporal.}
The three Theta variants (48--58\%) are competitive with the best LLM w/ tools.
Their strength is Lookup (up to 81.8\% for Theta Smart Expert), where
structured API calls directly address filter-and-retrieve operations.
However, Trend accuracy varies widely across Theta variants (35--70\%),
suggesting that the prompt design for temporal aggregation queries remains a
bottleneck even with structured access.

\paragraph{Memory RAG: weakest paradigm.}
HippoRAG consistently score lowest (30--38\%).
Comparison accuracy is particularly poor (2--18\%), because multi-hop joins
across events require evidence that is distributed across many chunks and
hard to consolidate through memory-based retrieval alone.
Increasing the retrieval budget $k$ from 10 to 50 does not consistently help:
HippoRAG improves on Comparison (7.7\% $\to$ 18.1\%) but not on other
dimensions.

\paragraph{Dimension-level patterns.}
Trend yields the highest accuracy for LLM w/ tools (71--95\%)---many queries
reduce to time-series aggregation that models handle well with tool support.
Anomaly is moderately easy (56--76\% for LLM/DB agents) since threshold
checks and abnormality lookups are straightforward.
Comparison difficulty varies sharply by paradigm: LLM and DB agents achieve
32--74\%, while Memory RAG collapses to 2--18\%.
Explanation proves universally hardest (15--31\% across all methods),
confirming that evidence-ranked attribution remains beyond current
architectures.

\paragraph{Easy--Medium--Hard gradient.}
The difficulty tier gradient is clear: Easy accuracy ranges from 56--76\%,
Medium from 23--72\%, and Hard from 22--55\%.
The steepest drops appear for LLM w/ tools on Hard queries (e.g., Sonnet 4.6:
73.1\% Easy $\to$ 37.9\% Hard), while DB agents degrade more gracefully
(Theta Expert: 73.8\% $\to$ 51.9\%).

\begin{table*}[t]
\centering
\caption{Main results by dimension: accuracy (\%). Methods are grouped by paradigm: LLM w/ tools (tool-use), DB agent (structured API), and Memory RAG. The base LLM for DB agent and Memory RAG methods is GPT-5.4. Total is the per-query average across all queries; dimension sub-scores are computed over varying query counts due to the sampling distribution.}
\label{tab:main_results_dim}
\begin{tabular}{l l r r r r r r}
\toprule
Method & Paradigm & Lookup & Trend & Comp. & Anom. & Expl. & Total \\
\midrule
Gemini 3 Flash Preview   & LLM w/ tools  & 73.5 & 94.8 & 74.4 & 64.3 & 25.3 & 62.9 \\
Theta General (GPT-5.4)  & DB agent   & 73.8 & 70.0 & 67.0 & 66.1 & 26.4 & 57.9 \\
Theta Expert (GPT-5.4)   & DB agent   & 75.6 & 49.5 & 68.1 & 71.2 & 30.9 & 56.4 \\
Sonnet 4.6               & LLM w/ tools  & 50.8 & 87.0 & 42.7 & 72.6 & 25.6 & 53.7 \\
MiniMax M2.5             & LLM w/ tools  & 43.1 & 87.0 & 43.7 & 73.4 & 17.5 & 49.6 \\
GPT-5.4                  & LLM w/ tools  & 52.3 & 76.5 & 32.0 & 75.7 & 17.9 & 49.2 \\
Theta Smart Expert       & DB agent   & 81.8 & 35.0 & 67.7 & 43.8 & 20.0 & 47.6 \\
GLM-5                    & LLM w/ tools  & 43.1 & 76.5 & 45.2 & 62.2 & 20.6 & 46.7 \\
GPT-5.2                  & LLM w/ tools  & 52.7 & 71.0 & 41.0 & 56.4 & 15.5 & 45.4 \\
HippoRAG ($k{=}50$)      & Memory RAG & 67.7 & 38.0 & 18.1 & 35.7 & 17.9 & 37.7 \\
HippoRAG ($k{=}10$)      & Memory RAG & 62.4 & 32.5 &  7.7 & 61.6 & 14.6 & 37.5 \\
HippoRAG ($k{=}20$)      & Memory RAG & 61.2 & 32.0 &  5.4 & 42.9 & 23.3 & 36.2 \\
\bottomrule
\end{tabular}
\end{table*}

\begin{table}[t]
\centering
\caption{Main results by difficulty tier: accuracy (\%). Total is the per-query average; tier sub-scores reflect the actual query distribution, which may deviate slightly from the 20/30/50 target split.}
\label{tab:main_results_tier}
\begin{tabular}{l r r r r}
\toprule
Method & Easy & Med. & Hard & Total \\
\midrule
Gemini 3 Flash Preview   & 75.0 & 70.6 & 55.0 & 62.9 \\
Theta General            & 76.2 & 61.3 & 50.5 & 57.9 \\
Theta Expert             & 73.8 & 55.2 & 51.9 & 56.4 \\
Sonnet 4.6               & 73.1 & 71.6 & 37.9 & 53.7 \\
MiniMax M2.5             & 56.2 & 67.0 & 38.0 & 49.6 \\
GPT-5.4                  & 73.8 & 60.7 & 35.6 & 49.2 \\
Theta Smart Expert       & 65.6 & 45.0 & 43.8 & 47.6 \\
GLM-5                    & 55.6 & 72.2 & 29.9 & 46.7 \\
GPT-5.2                  & 65.0 & 64.2 & 29.1 & 45.4 \\
HippoRAG ($k{=}50$)      & 68.8 & 34.3 & 30.4 & 37.7 \\
HippoRAG ($k{=}10$)      & 57.3 & 34.9 & 33.0 & 37.5 \\
HippoRAG ($k{=}20$)      & 70.0 & 23.1 & 33.6 & 36.2 \\
\bottomrule
\end{tabular}
\end{table}

\subsection{Error analysis}
\label{sec:error-analysis}

Where do methods break? We manually inspect incorrect responses sampled from
Medium and Hard tiers across paradigms. Two representative failure cases
follow.

\paragraph{Case 1: Cross-source indicator confusion (HippoRAG,
Comparison/Hard).}
The query asks which events share an affected indicator with a given event.
The ground truth requires joining through the indicator \emph{systolic blood
pressure}, which appears in both device and exam artifacts.
HippoRAG retrieves the correct event chunk but also retrieves an exam chunk
referencing \emph{diastolic blood pressure}, leading the LLM to include a
spurious event in the answer set.
The root cause is indicator-level disambiguation failure: chunk boundaries
split related fields, and semantic similarity alone cannot resolve the
ambiguity.

\paragraph{Case 2: Temporal window misalignment (GPT-5.2,
Comparison/Medium).}
The query asks for the mean resting heart rate during the 14-day pre-event
window of a specific event compared with the during-event period.
GPT-5.2 identifies the correct event but miscalculates the window boundaries,
using the event \emph{end} date instead of the \emph{start} date as the
reference point.
As a result, the pre-event window overlaps with the active event period,
inflating the reported mean.
This type of temporal anchoring error is systematic across LLM w/ tools methods
and accounts for a substantial fraction of Comparison and Lookup failures at
the Medium and Hard tiers.

\section{Discussion and Limitations}
\label{sec:discussion}

\paragraph{\dataset{} as an evaluation target.}
Longitudinal reasoning skills---temporal alignment across sources, multi-hop
joins over event--indicator structure, window statistics,
evidence-structured attribution---often fail silently in open-ended demos.
Verifiable ground truth makes them measurable; the dimension--tier taxonomy
then pinpoints which capabilities a given architecture lacks.

\paragraph{Generalizability.}
Strong performance on \dataset{} does not guarantee equivalent accuracy on
real EHR data: our generator produces cleaner temporal structure and more
regular event boundaries than typical clinical records.
We view \dataset{} as a necessary-condition test: a method that fails
multi-hop joins and temporal windowing under these idealized conditions is
unlikely to succeed on noisier real-world data.
Calibrating generator priors against real-cohort summary statistics---event
frequency, indicator distributions---is a concrete path to narrow the
external-validity gap.

\paragraph{Limitations.}
\dataset{} is not a physiological simulator. Event-to-indicator dynamics are
simplified; effect magnitudes should not be read as clinical effect sizes.
Representativeness hinges on profile priors, event catalogs, and prompt
templates---all of which can introduce biases from developer assumptions or
from the LLM's training data.
LLM-generated narratives may drift across model versions, so reproducibility
requires fixed seeds, versioned prompts, and explicit configuration files.
The Explanation dimension evaluates mechanism recovery under known generator
dynamics, not causal inference in observational medicine.

\paragraph{Generation cost.}
All synthetic data is generated with Gemini 3.1 Pro Preview
(\texttt{gemini-3.1-pro-preview-thinking}).
A single user averages 614 API calls, consuming ${\sim}$20.3M input tokens
and ${\sim}$2.8M output tokens---roughly \$74 at current Gemini Pro rates
(\$2.00/1M input, \$12.00/1M output including thinking tokens).
Event decision and impact generation dominate the cost; however, the sparsity
gate (\cref{sec:framework}) suppresses event decisions on most days, so the
average call count (${\sim}$614) is well below the observation span
(${\sim}$1{,}196~days).
Device trajectory simulation is purely algorithmic and adds negligible
overhead.

\paragraph{Responsible use.}
Synthetic does not mean safe. Membership inference attacks have been
demonstrated against synthetic health data under realistic threat
models~\cite{zhang2022mia}.
Safeguards should include clear labeling to prevent mixing with real patient
data, synthetic identifiers in place of personal information, documented
generation assumptions in accompanying data cards, and a prohibition on
using systems evaluated on \dataset{} for medical diagnosis or clinical
advice.

\paragraph{Extensions.}
Several directions are worth pursuing: modeling irregular sampling and device
non-wear as first-class processes; richer cross-indicator constraints beyond
shared noise factors; partial-credit metrics (set-F1, tolerance sweeps);
multilingual query support; and real-data calibration that uses cohort
summary statistics to anchor event frequencies and indicator distributions.
Incorporating additional modalities---imaging reports, free-text clinical
notes---would further broaden the benchmark's scope.

\section{Conclusion}
\label{sec:conclusion}

\dataset{} is an event-driven benchmark for longitudinal health agents.
Patient trajectories are modeled as baseline health states superposed with
discrete events whose temporal kernels---sigmoid onset, exponential
fade-out---are fully specified and verifiable; a multi-phase trajectory plan
ensures longitudinal narrative coherence.
One hundred synthetic users, spanning 1--5 years of daily device streams,
sparse exams, and structured event logs, are paired with 10{,}000 queries
across five dimensions and three difficulty tiers under a two-stage scoring
protocol.

The empirical picture is consistent: DB agents (48--58\%) substantially
outperform memory RAG baselines (30--38\%), with the gap concentrated on
Comparison and Explanation queries where multi-hop reasoning and evidence
attribution are required.
The bottleneck is structured temporal reasoning---cross-source joins,
event-aligned windowing, evidence-organized attribution---not language
understanding.
If \dataset{} can help steer agent development toward these temporal
reasoning capabilities, it will have served its purpose.

\subsection*{Declaration of AI Use}

We acknowledge the use of AI-assisted technologies in the preparation of this manuscript:

\begin{itemize}
\item \textbf{Writing Assistance:} We employed AI language models (Claude Opus 4.6 and GPT-5.2 Pro) to assist with drafting, editing, and refining the clarity, grammar, and structure of the text. All scientific arguments, experimental design, data analysis, and conclusions were conceived, verified, and approved by the authors.

\item \textbf{Code Development:} AI tools were used to assist with implementing the synthesis framework, evaluation query generators, and analysis scripts. All code was reviewed, tested, and validated by the authors.

\item \textbf{Figure Generation:} Architecture diagrams and conceptual figures were generated with the assistance of Nano Banana Pro, then reviewed and adjusted by the authors to ensure accuracy.

\item \textbf{Synthetic Data Generation:} The longitudinal health data in \dataset{} was generated using Gemini 3.1 Pro Preview as the LLM backbone for agentic components (profile generation, event decisions, exam generation). The algorithmic simulation components (indicator dynamics, constraint enforcement) are deterministic and do not involve AI generation.
\end{itemize}

\noindent The authors take full responsibility for the accuracy and integrity of all content in this work.

\appendix
\section{Agent Prompt Templates}
\label{app:prompts}

This appendix summarizes the core LLM prompts used in the \dataset{} synthesis
pipeline and the evaluation rubric used for scoring.
Each prompt is shown in abbreviated form; full prompts with examples, rubric
definitions, and the judge prompt are available in the codebase.

\subsection{Profile Generation}

\begin{promptblock}
Role: Senior clinical physician and health profile expert.

Input: Demographics JSON, personality narrative.

Task: Generate a complete, realistic health profile.

Guiding principles:
1) Personality-Health: reflect traits in lifestyle and mental health.
2) Occupation-Health: match occupational risks to posture, activity, stress.
3) Age-Health: health issues must match age characteristics.
4) Family history: 30-50% have none; generate only when clinically motivated.
5) Physical measurements: height (cm), weight (kg), BMI 17-40 (accommodating obesity profiles).

Output: Structured JSON (demographics, physical_measurements, health_profile,
lifestyle, family_history, medication_profile, user_narrative).
\end{promptblock}

\subsection{Indicator Selection}

\begin{promptblock}
Role: Medical healthcare expert for indicator selection.

Input: User profile, available device/exam indicator catalog, target counts.

Task: Select the most relevant indicators via two-stage reasoning.

Stage 1 - Analysis: user characteristics, risk assessment, priority strategy.
Stage 2 - Selection: device (daily: HR, BP, steps, sleep) and exam (periodic
labs, organ-function panels), each with per-indicator selection_reason.

Output: Structured JSON (selection_reasoning, selected_device_indicators,
selected_exam_indicators).
\end{promptblock}

\subsection{Trajectory Planning}

\begin{promptblock}
Role: Medical expert and narrative designer.

Input: User profile, observation date range, total days, initial events.

Task: Produce a phased health trajectory plan (the "director's script") for
the entire observation period.

Constraints:
1) Each phase covers approximately 3-5 months; no gaps or overlaps.
2) Each phase has a unique health theme and at least one turning point.
3) The overall arc must have realistic ups and downs -- not monotonically
   improving.
4) Use directional descriptions only (no specific indicator values).
5) Chronic-disease users: tell a disease-management story (exacerbations,
   treatment adjustments, recovery arcs).
6) Healthy users: tell a lifestyle-driven story (fitness changes, stress
   cycles, weight fluctuations, acute episodes).

Output: Structured JSON (overall_narrative, planning_reasoning,
phases[{phase_number, phase_name, start_date, end_date, description}]).
\end{promptblock}

\subsection{Event Decision}

\begin{promptblock}
Role: Medical expert for daily event generation decisions.

Input: User profile, current date, observation progress, recent 3-day device
data, recent exam abnormals, historical events, diversity report, trajectory
phase context, active events.

Task: Decide whether to generate a new health event on this day.

Trajectory alignment (when trajectory is available):
1) Read the current phase description -- this is the storyline for this period.
2) Storyline events (trajectory-aligned): events that directly reflect the
   phase theme. These receive highest priority.
3) Texture events (background): everyday occurrences (seasonal colds, social
   dinners, minor injuries) that do not contradict the trajectory. Welcome for
   realism and diversity.
4) Trajectory gap check: if the phase describes a specific change that no
   recent event has realized, prioritize filling that gap.
5) Only avoid events that directly contradict the phase direction.

Decision process (4 stages):
Stage 1: Candidate identification across 7 dimensions (indicator-driven,
family history, occupation, habit, sudden, chronic disease, habit recovery).
Stage 2: Coverage check (no duplicates with active events).
Stage 3: Reasonability check (feasibility, plausibility, burden/conflict).
Stage 4: Priority selection (urgency, impact scope, trajectory alignment).

Event types: diet_change, exercise_change, health_event, long_term_habit.

Output: Structured JSON (analysis_reasoning, should_generate, event_details
with event_type, event_name, description, duration_days, health_effect,
trigger_reason, interrupts_events).
\end{promptblock}

\subsection{Event Indicator Impact}

\begin{promptblock}
Role: Medical expert for event-to-indicator impact quantification.

Input: User profile, current date, event details, available device/exam
indicators with baseline values and units.

Task: Generate per-indicator impact parameters via chain-of-thought reasoning.

Reasoning chain:
1) Event nature and primary physiological systems affected.
2) Cascade: primary -> secondary -> longer-term adaptations.
3) Per-indicator: affected? direction? magnitude? onset? fade-out?
4) Quantification (expected_change in the indicator's native unit).
5) Validity checks (medical, magnitude, temporal plausibility).

Output: Structured JSON (impact_reasoning, affected_indicators list with
indicator_name, expected_change, impact_level, time_to_effect, fade_out_days,
influence_reasoning).
\end{promptblock}

\bibliographystyle{unsrt}
\bibliography{refs}

\end{document}